\documentclass[10pt,journal,compsoc]{IEEEtran}

\usepackage[pagebackref=true,breaklinks=true,letterpaper=true,colorlinks,bookmarks=false]{hyperref}
\ifCLASSOPTIONcompsoc
  \usepackage[nocompress]{cite}
\else
  \usepackage{cite}
\fi
\ifCLASSINFOpdf
\else
\fi
\usepackage{times}
\usepackage{epsfig}
\usepackage{graphicx}
\usepackage{amsmath}
\usepackage{amssymb}
\usepackage{subcaption} 
\usepackage{multirow}
\usepackage{rotating}
\usepackage{booktabs}
\usepackage{pifont}
\usepackage[inline]{enumitem}
\usepackage{enumitem}
 \usepackage{array,multirow}

\usepackage{comment}

\newcommand{\xmark}{\ding{55}}%

\usepackage{algpseudocode}
\usepackage{algorithm}

\algnewcommand\algorithmicforeach{\textbf{for each}}
\algdef{S}[FOR]{ForEach}[1]{\algorithmicforeach\ #1\ \algorithmicdo}

\usepackage{color}

\begin{document}
%
\title{Depthwise Spatio-Temporal STFT Convolutional Neural Networks for Human Action Recognition}
%
%
%
%

\author{Sudhakar~Kumawat, 
	Manisha~Verma, 
	Yuta~Nakashima, 
	and~Shanmuganathan~Raman 
\IEEEcompsocitemizethanks{\IEEEcompsocthanksitem Sudhakar Kumawat is  with 
Computer Science and Engineering, Indian Institute of Technology Gandhinagar, Gujarat, India, 382355. E-mail: sudhakar.kumawat@iitgn.ac.in
\IEEEcompsocthanksitem Manisha Verma and Yuta Nakashima are with the Intelligence and Sensing Lab, Institute for Datability Science, Osaka University, Osaka, Japan, 567-0871. E-mail:  \{mverma, n-yuta\}@ids.osaka-u.ac.jp
\IEEEcompsocthanksitem Shanmuganathan Raman is  with  Electrical Engineering and Computer Science and Engineering, Indian Institute of Technology Gandhinagar, Gujarat, India, 382355. E-mail: shanmuga@iitgn.ac.in
}
}

\IEEEtitleabstractindextext{%
\begin{abstract}
Conventional 3D convolutional neural networks (CNNs) are computationally expensive, memory intensive, prone to overfitting, and most importantly, there is a need to improve their feature learning capabilities. To address these issues, we propose spatio-temporal short term Fourier transform (STFT) blocks, a new class of convolutional blocks that can serve as an alternative to the 3D convolutional layer and its variants in 3D CNNs. An STFT block consists of non-trainable convolution layers that capture spatially and/or temporally local Fourier information using a  STFT kernel at multiple low frequency points, followed by a set of trainable linear weights for learning channel correlations. The STFT blocks significantly reduce the space-time complexity in 3D CNNs. In general, they use 3.5 to 4.5 times less parameters and 1.5 to 1.8 times less computational costs when compared to the state-of-the-art methods. Furthermore, 
their feature learning capabilities are significantly better than the conventional 3D convolutional layer and its variants. Our extensive evaluation on seven action recognition datasets, including Something$^2$ v1 and v2, Jester,  Diving-48, Kinetics-400, UCF 101, and HMDB 51, demonstrate that STFT blocks based 3D CNNs achieve on par or even better performance compared to the state-of-the-art methods.

 
\end{abstract}

}

\maketitle

\IEEEdisplaynontitleabstractindextext

%
\IEEEpeerreviewmaketitle

\IEEEraisesectionheading{\section{Introduction}\label{sec:introduction}}

In recent years, with the availability of large-scale datasets and computational power, deep neural networks (DNNs) have led to unprecedented advancements in the field of artificial intelligence. In particular, in computer vision, research in the area of convolutional neural networks (CNNs) has achieved impressive results on a wide range of applications such as robotics \cite{kumra2017robotic}, autonomous driving \cite{wu2017squeezedet}, medical imaging \cite{litjens2017survey}, face recognition \cite{masi2018deep}, and many more. This is especially true for the case of 2D CNNs where they have achieved unparalleled performance boosts on various computer vision tasks such as image classification \cite{he2016deep,iandola2016squeezenet}, semantic segmentation \cite{chen2017deeplab}, and object detection \cite{ren2015faster}.

Unfortunately, 3D CNNs, unlike their 2D counterparts, have not enjoyed a similar level of performance jumps on tasks that require to model spatio-temporal information, e.g.~video classification. In order to reduce this gap, many attempts have been made by developing bigger and challenging datasets for tasks such as action recognition. However, many challenges still lie in the architecture of deep 3D CNNs. Recent works, such as \cite{xie2018rethinking} and \cite{hara2018can}, have listed some of the fundamental barriers in designing and training deep 3D CNNs due to a large number of parameters, such as:
\begin{enumerate*}[label=(\arabic*), ref=\arabic*]
	\item they are computationally very expensive,
	\item they result in a large model size, both in terms of memory usage and disk space,
	\item they are prone to overfitting, and
	\item very importantly, there is a need to improve their capability to capture spatio-temporal features, which may require fundamental changes in their architectures \cite{xie2018rethinking,hara2018can,citation-0,tran2018closer}.
\end{enumerate*}
Some of them are common in 2D CNNs as well for which various techniques have been proposed in order to overcome these barriers \cite{howard2017mobilenets,zhang2017shufflenet}. However, direct transfer of these techniques to 3D CNNs falls short of achieving expected performance goals \cite{kopuklu2019resource}. 
\begin{figure*}[t]
	\centering
	
	\fbox{	
		\subcaptionbox{\label{fig:a}}{\includegraphics[width=.13\columnwidth, height=0.45\columnwidth]{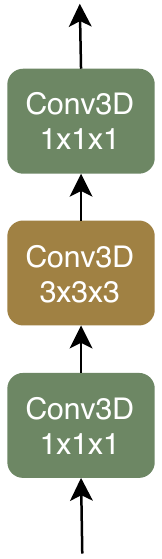}}\hspace{1em}%
		\subcaptionbox{\label{fig:b}}{\includegraphics[width=.2\columnwidth, height=0.45\columnwidth]{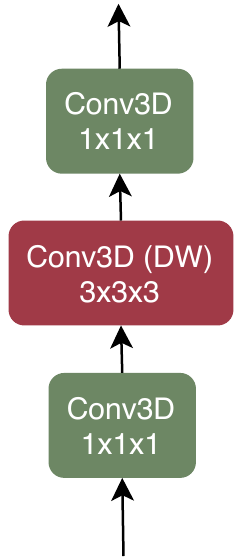}}\hspace{1em}%
		\subcaptionbox{\label{fig:c}}{\includegraphics[width=.2\columnwidth, height=0.45\columnwidth]{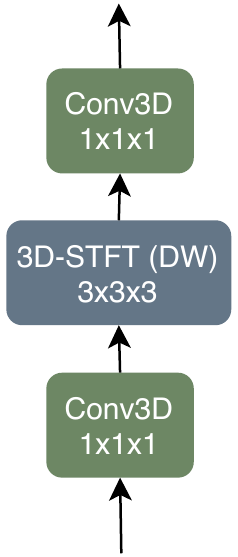}}}\hspace{3em}%
	\fbox{
		\subcaptionbox{\label{fig:d}}{\includegraphics[width=.13\columnwidth, height=0.5\columnwidth]{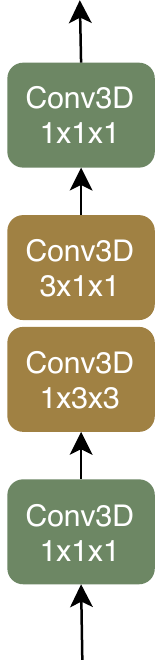}}\hspace{1em}%
		\subcaptionbox{\label{fig:e}}{\includegraphics[width=.2\columnwidth, height=0.5\columnwidth]{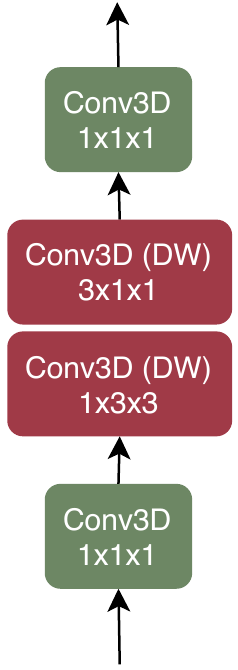}}\hspace{1em}%
		\subcaptionbox{\label{fig:f}}{\includegraphics[width=.2\columnwidth, height=0.5\columnwidth]{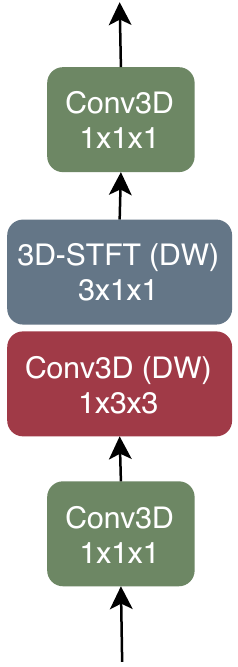}}\hspace{1em}%
		\subcaptionbox{\label{fig:g}}{\includegraphics[width=.2\columnwidth, height=0.5\columnwidth]{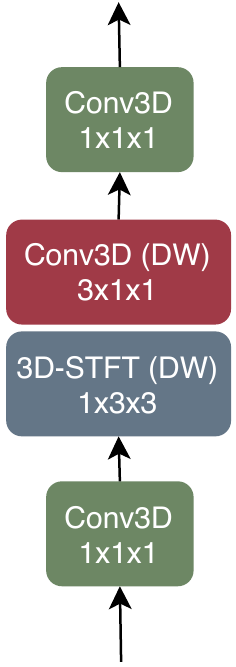}}
		}
	\caption{Illustration of bottleneck versions of various 3D convolutional layers and its variations. (a) Standard 3D convolutional layer-based  block used in I3D \cite{carreira2017quo}. (b) Standard 3D depthwise (DW) convolutional layer-based  block used in CSN \cite{tran2019video}. (c) The ST-STFT block. (d) Factorized 3D convolutional layer-based block used in S3D \cite{xie2018rethinking}. (e) Factorized + depthwise 3D convolutional layer-based  block (f) The T-STFT block. (g) The S-STFT block. Note that here instead of using 3D-STFT, 2D-STFT or 1D-STFT, we will use a common notation 3D-STFT for the STFT kernels in all the three variations of the STFT blocks. The dimension(s) of the information captured by the STFT block will be denoted by the filter size. DW denotes depthwise.}
	\label{fig:layer_types}
\end{figure*}
 

The primary source of high computational complexity in 3D CNNs is the traditional 3D convolutional layer itself. The standard implementation of the 3D convolutional layer learns the spatial, temporal, and channel correlations simultaneously. 
This leads to very dense connections that leads to high complexity and accompanying issues, such as overfitting. Recent methods attempt to address this problem by proposing efficient alternatives to the 3D convolutional layer. 

For example, R(2+1)D \cite{tran2018closer}, S3D \cite{xie2018rethinking}, and P3D \cite{qiu2017learning} factorize 3D convolutional kernels into two parts, one for spatial dimensions and the other for temporal dimension. 3D versions \cite{kopuklu2019resource} of MobileNet \cite{howard2017mobilenets} and ShuffleNet \cite{zhang2017shufflenet} factorize the kernel along the channel dimension, separately learning the channel dependency and the spatio-temporal dependency. The former uses \textit{pointwise} convolutions, of which kernels cover only the channel dimension, whereas the latter uses  \textit{depthwise}\footnote{This term may be confusing in the context of 3D CNNs. Following the convention, we use \textit{depthwise} to denote a 3D convolution that only covers the spatio-temporal dimensions and shared among the channel dimension.} 3D convolutions, of which kernel covers the spatio-temporal dimensions. Similarly, CSN \cite{tran2019video} uses depthwise or \textit{group} convolutions \cite{krizhevsky2012imagenet} along with pointwise \textit{bottleneck}\footnote{We use \textit{bottleneck} to refer to the bottleneck architecture in ResNet \cite{he2016deep}.} layers in order to separate the learning of channel and spatio-temporal dependencies. Figure \ref{fig:layer_types} illustrates various convolutional layers blocks built on top of the bottleneck architecture, which cover all input dimensions; a standard convolution layer in Figure~\ref{fig:a}, a depthwise 3D convolution layer (e.g.~CSN) in Figure~\ref{fig:b}, and a factorized 3D convolution (e.g.~S3D) in Figure~\ref{fig:d}. 

Other works such as STM \cite{jiang2019stm}, MiCT \cite{zhou2018mict}, and GST \cite{luo2019grouped} propose to augment state-of-the-art 2D CNNs with temporal aggregation modules in order to efficiently capture the spatio-temporal features and leverage the relatively lower complexity of 2D CNNs. However, despite this impressive progress of deep 3D CNNs on human action recognition, their complexity still remains high in comparison to their 2D counterparts. This calls for the need to develop resource efficient 3D CNNs for real-time applications while taking in account their runtime, memory, and power budget.

In this work, we introduce a new class of convolution blocks, called spatio-temporal short term Fourier transform (STFT) blocks, that serve as an alternative to traditional 3D convolutional layer and its variants in 3D CNNs. The STFT blocks broadly comprises of a depthwise STFT layer, which uses a \emph{non-trainable} STFT kernel, and a set of \emph{trainable} linear weights. The depthwise STFT layer extracts depthwise local Fourier coefficients by computing STFT \cite{hinman1984short} at multiple low frequency points in a local $n_t\times n_h\times n_w$ (e.g., $3\times 3\times 3$) volume of each position of the input feature map. 
The output from the depthwise STFT layer is then passed through a set of trainable linear weights that computes weighted combinations of these feature maps to capture the inter-channel correlations. 

We propose three variants of STFT blocks depending on the dimensions that STFT kernels cover. In the \emph{first} variant, as shown in Figure~\ref{fig:c}, both the spatial and temporal information are captured using a non-trainable spatio-temporal 3D STFT kernel. We call this block as ST-STFT. In the \emph{second} variant, which we call S-STFT (Figure~\ref{fig:f}), a non-trainable spatial 2D STFT kernel captures only the spatial information and a trainable depthwise 3D convolution captures the temporal information. In the \emph{third} variant in Figure~\ref{fig:g}, the spatial information is captured using the trainable depthwise 3D convolutions, and temporal 1D STFT kernel captures the temporal information. We call this block T-STFT.

Our proposed STFT blocks provide significant reduction of the number of parameters along with computational and memory savings. STFT block-based 3D CNNs have much lower complexity and are less prone to overfitting. Most importantly, their feature learning capabilities (both spatial and temporal) are significantly better than conventional 3D convolutional layers and their variants. In summary, the main contributions of this paper are as follows:

\begin{itemize}
\item We propose depthwise STFT layers, a new class of 3D convolutional layers that uses STFT kernels to capture spatio-temporal correlations. We also propose STFT blocks, consisting of a depthwise STFT layer and some trainable linear weights, which can replace traditional 3D convolutional layers.

\item We demonstrate that STFT block-based 3D CNNs consistently outperform or attain a comparable performance with the state-of-the-art methods on seven publicly available benchmark action recognition datasets, including Something$^2$ V1 \& V2 \cite{goyal2017something}, Diving-48
\cite{li2018resound}, Jester \cite{yogajournal}, Kinetics \cite{carreira2017quo}, UCF101 \cite{soomro2012ucf101}, and HMDB-51 \cite{kuehne2011hmdb}.

\item We show that the STFT blocks significantly reduce the complexity in 3D CNNs. In general, they use 3.5 to 4.5 times less parameters and 1.5 to 1.8 times less computations when compared with the state-of-the-art methods. 

\item We present detailed ablation and performance studies
for the proposed STFT blocks. This analysis will be useful for exploring STFT block-based 3D CNNs in future.
\end{itemize}

A preliminary version of this work was published in CVPR 2019 \cite{kumawat2019lp}. Compared to the conference version, we substantially extend and improve the following aspects.
\begin{enumerate*}[label=(\roman*)]
	\item In Section~\ref{sec:variations}, we introduce an improved version (i.e., ST-STFT) of the ReLPV block of \cite{kumawat2019lp} by integrating the concept of depthwise convolutions into the STFT layer of ReLPV; thus, significantly improving its performance.  
	\item Additionally, in Section~\ref{sec:variations}, we propose two new variants of ST-STFT, referred to as S-STFT and T-STFT. These new variants differ in the manner the STFT kernel capture either only spatial or temporal dependencies, respectively.
	\item In Section \ref{sec:STFTnets}, we introduce new network architectures based on the ST-STFT, S-STFT, and T-STFT blocks. 
	\item In Section \ref{sec:ablation_studies}, we conduct extensive ablation and performance studies for the proposed STFT blocks.  
	\item In Section~\ref{sec:results_AR}, we present extensive evaluation of the STFT block-based CNNs on seven action recognition datasets and compare with the state-of-the-art methods. \item Finally, we extend Section~\ref{sec:related_work} and provide an extensive literature survey of various  state-of-the-art methods on human action recognition. Furthermore, in Section~\ref{sec:derivation_stft}, we provide detailed mathematical formulation and visualization for the STFT layer. 
\end{enumerate*}

The rest of the paper is organized as follows. Section \ref{sec:related_work} extensively reviews the related literature on human action recognition. Section \ref{sec:method} introduces our STFT blocks. Section \ref{sec:STFTnets} illustrates the architecture of the STFT block-based networks. Section \ref{sec:experiments} discusses various action recognition datasets used for evaluation and implementation details.   Section \ref{sec:ablation_studies} provides detailed ablation and performance studies of the STFT block-based networks. Section \ref{sec:results_AR} presents experimental results and comparisons with state-of-the-art methods on benchmark action recognition datasets. Finally, Section \ref{sec:conclusion} concludes the paper with future directions.

\section{Related Work}\label{sec:related_work}
In recent years, deep convolutional neural networks have accomplished unprecedented success on the task of object recognition in images \cite{he2016deep, xie2017aggregated,hu2018squeeze}. Therefore, not surprisingly, there have been many recent attempts to extend this success to the domain of human action recognition in videos \cite{karpathy2014large,simonyan2014two}. Among the first such attempts, Karpathy \emph{et al.} \cite{karpathy2014large} applied 2D CNNs on each frame of the video independently and explored several approaches for fusing information along the temporal dimension. However, the method achieved inferior performance as the temporal fusion fell short in terms of modeling the interactions among frames.

\textbf{Optical flow-based methods.} For better temporal reasoning, Simonyan and Zissernan \cite{simonyan2014two} proposed a two-stream CNN architecture where the first stream, called a spatial 2D CNN, would learn scene information from still RGB frames and the second stream, called a temporal 2D CNN, would learn temporal information from pre-computed optical flow frames. Both the streams are trained separately and the final prediction for the video is averaged over the two streams. Several works such as \cite{feichtenhofer2017spatiotemporal,feichtenhofer2016convolutional,wang2016temporal,wang2017spatiotemporal} further explored this idea. For example, Feichtenhofer \emph{et al.} \cite{christoph2016spatiotemporal,feichtenhofer2017spatiotemporal} proposed fusion strategies between the two streams in order to better capture the spatio-temporal features. Wang \emph{et al.} \cite{wang2017spatiotemporal} proposed a novel spatio-temporal pyramid network to fuse the spatial and temporal features from the two streams. Wang \emph{et al.} proposed temporal segment networks (TSNs) \cite{wang2016temporal}, which utilizes a sparse temporal sampling method and fuses both the streams using a weighted average in the end. 
A major drawback of the two-stream networks is that they require to pre-compute optical flow, which is expensive in terms of both time and space \cite{zach2007duality}. 

\textbf{Conventional 3D convolutional layer-based methods.} In order to avoid the complexity associated with the optical flow-based methods, Tran \emph{et al.} \cite{tran2015learning} proposed to learn the spatio-temporal correlations directly from the RGB frames by using 3D CNNs with standard 3D convolutional layers. Later, Carreira and Zisserman proposed I3D \cite{carreira2017quo} by inflating 2D kernels of GoogleNet architecture \cite{szegedy2015going} pre-trained on ImageNet  into 3D in order to efficiently capture the spatio-temporal features. Other works, such as \cite{hara2017learning,tran2017convnet} used the idea of residual blocks from ResNet architectures \cite{he2016deep} to improve the performance of 3D CNNs. However, due to various constraints, such as lack of large scale action recognition datasets and large number of parameters associated with 3D convolutional layer, all the above works explored shallow 3D CNN architectures only. Hara \emph{et al.} \cite{hara2018can} undertook the first study of examining the performance of deep 3D CNN architectures on a large scale action recognition dataset. They replaced 2D convolution kernels with their 3D variants in deep architectures such as ResNet \cite{he2016deep} and ResNext \cite{xie2017aggregated} and provided benchmark results on Kinetics \cite{carreira2017quo}.

Inspired by the performance of SE (Squeeze and Excitation) blocks of SENet \cite{hu2018squeeze} on ImageNet, Diba \emph{et al.} \cite{diba2018spatio} proposed a spatio-temporal channel correlation (STC) block that can be inserted into any 3D ResNet-style network to model the channel correlations among spatial and temporal features throughout the network. Similarly, Chen \emph{et al.} proposed a double attention block \cite{chen20182} for gathering and distributing long-range spatio-temporal features. Recently, Feichtenhofer \emph{et al.} proposed SlowFast network \cite{feichtenhofer2019slowfast} that use a slow pathway, operating at a low frame rate, to capture the spatial correlations and a fast pathway, operating at a high frame rate, to capture motion at a fine temporal resolution. More recently, Stroud \emph{et al.} introduced the distilled 3D network (D3D) architecture \cite{stroud2020d3d} that uses knowledge distillation from the temporal stream to improve the motion representations in the spatial stream of 3D CNNs.

\textbf{Factorized 3D convolutional layer-based methods.} As mentioned earlier,  standard 3D convolutional layers simultaneously capture the spatial, temporal, and channel correlations, which lead to a high complexity. Various methods proposed to solve this problem by factorizing this process along different dimensions \cite{tran2018closer,xie2018rethinking,qiu2017learning,kopuklu2019resource,tran2019video}. For example, R(2+1)D \cite{tran2018closer}, S3D \cite{xie2018rethinking}, and P3D \cite{qiu2017learning} factorize a 3D convolutional layer into a 2D spatial convolution followed by a 1D temporal convolution, separating the kernel to cover the channel$+$spatial and channel$+$temporal dimensions. Kopuklu \emph{et al.} \cite{kopuklu2019resource} studied the effect of such factorization along the channel dimension by extending MobileNet \cite{howard2017mobilenets} and ShuffleNet \cite{zhang2017shufflenet} architectures into 3D. Similarly, Tran \emph{et al.} in CSN \cite{tran2019video} explored bottleneck architectures with group \cite{krizhevsky2012imagenet} and depthwise 3D convolutions, which factorizes a 3D convolutional layer along the channel dimension. These works showed that factorizing 3D convolutional layers provides a form of regularization that leads to an improved accuracy in addition to a lower computational cost.

\textbf{Temporal aggregation modules.} Some efforts have also been dedicated to learn the temporal correlations using conventional 2D CNNs, which are computationally inexpensive, with using various temporal aggregation modules \cite{zhou2018mict,luo2019grouped,lin2019tsm,jiang2019stm, sun2018optical}. For example, MiCT \cite{zhou2018mict} and GST \cite{luo2019grouped} integrate 2D CNNs using 3D convolution layers for learning the spatio-temporal information. Lin \emph{et al.} \cite{lin2019tsm} introduced the temporal shift module (TSM) that shifts some parts of the channels along the temporal dimension, facilitating exchange of information among neighboring frames. Lee \emph{et al.} in MFNet \cite{lee2018motion} proposed fixed motion blocks for encoding spatio-temporal information among adjacent frames in any 2D CNN architectures. Jiang \emph{et al.} \cite{jiang2019stm} proposed the STM block that can be inserted into any 2D ResNet architecture. The STM block includes a depthwise spatio-temporal to obtain the spatio-temporal features and a channel-wise motion module to efficiently encode motion features. Similarly, Sun \emph{et al.} \cite{sun2018optical} proposed the optical flow-guided feature (OFF) that uses a set of classic operators, such as Sobel and element-wise subtraction, for generating OFF which captures spatio-temporal information. The main task of the temporal aggregation modules discussed in the above works is to provide an alternative motion representation while leveraging the computation and memory gains afforded by 2D CNN architectures. 

\begin{figure*}[!t]
	\begin{center}
		\includegraphics[width=\textwidth]{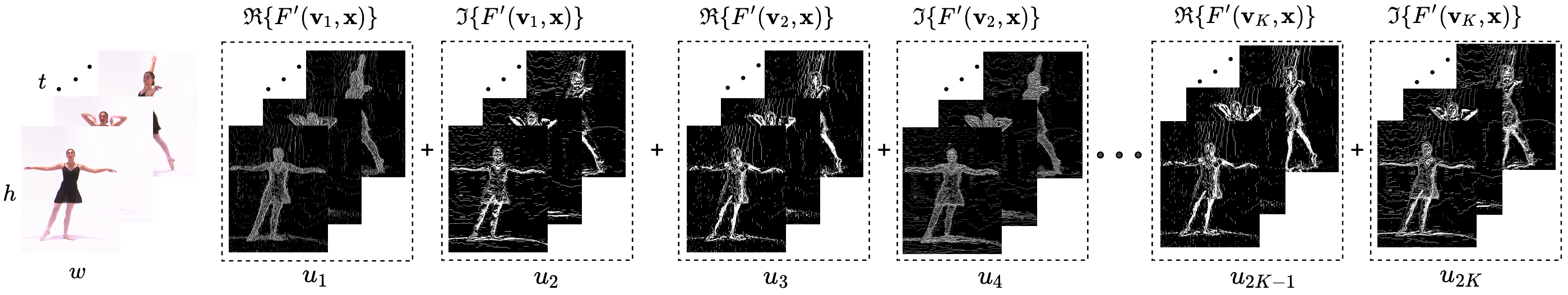}
		\caption{A Visualization of the output of the 3D-STFT layer applied to the input layer for $c = 1$, $n = 3$, and $K=13$. First, the Fourier coefficients are extracted in the local $3 \times 3 \times 3$ neighborhood of each pixel position of the input (at 13 frequency variables) to output a feature map of size $26\times t\times h \times w$. The output feature maps are then linearly combined using weights $u_1$,\ldots,$u_{13}$ and fed into the next layer.  Note that the STFT layer can be applied to any intermediate feature map in 3D CNN. Here, for simplicity, we visualize it on the input (first) layer. }
		\label{fig:depthwise3DSTFT}
	\end{center}
\end{figure*}

\section{STFT Blocks}\label{sec:method}

An STFT block consists of a non-trainable depthwise STFT layer, sandwiched by trainable pointwise convolutional layers. Note that, there are three variants of an STFT block as mentioned in Section~\ref{sec:introduction}; however, in what follows, we only provide the mathematical formulation of the ST-STFT block. The corresponding formulations for the S-STFT and T-STFT blocks follow exactly the same idea except the shape of kernels (i.e., $n_t \times n_h \times n_w$ for ST-STFT, $1 \times n_h \times n_w$ for S-STFT, and $n_t \times 1 \times 1$ for T-STFT). We discuss the details of the other variants in Section~\ref{sec:variations}.

\textbf{Why STFT for feature learning?} STFT in a multidimensional space was first explored by Hinman \emph{et al.} in \cite{hinman1984short} as an efficient tool for image encoding. It has two important properties which makes it useful for our purpose: (1)  Natural images are often composed of objects with sharp edges. It has been observed that Fourier coefficients accurately represent these edge features. Since STFT in the 3D space simply is a windowed Fourier transform, the same property applies \cite{hinman1984short}. Thus, STFT has the ability to accurately capture the local features in the same way as done by the convolutional filters. (2) STFT decorrelates the input signal \cite{hinman1984short}. Regularization is the key for deep learning as it allows training of more complex models while keeping lower levels of overfitting and achieves better generalization. Decorrelation of features, representations, and hidden activations has been an active area of research for better generalization of DNNs, with a variety of novel regularizers being proposed, such as DeCov \cite{cogswell2015reducing}, decorrelated batch normalization (DBN) \cite{DBLP:journals/corr/abs-1804-08450}, structured decorrelation constraint (SDC) \cite{xiong2016regularizing} and OrthoReg \cite{rodriguez2016regularizing}.

\subsection{Derivation of STFT kernels}\label{sec:derivation_stft}

We denote the feature map output by a certain layer in a 3D CNN network by tensor $f \in \mathbb{R}^{c\times t \times h \times w}$,  where $c$, $t$, $h$, and $w$ are the numbers of channels, frames, height, and width of $f$, respectively. For simplicity, let us take $c = 1$; hence, we can drop the channel dimension and rewrite the size of $f$ to $t\times h\times w$. We also denote a single element in $f$ by $f(\mathbf{x})$, where $\mathbf{x}\in\mathbb{Z}^3$ is the 3D (spatio-temporal) coordinates of element $f(\mathbf{x})$.

\begin{figure}[t]
	\begin{center}
		\includegraphics[width=0.7\columnwidth, height=0.5\columnwidth]{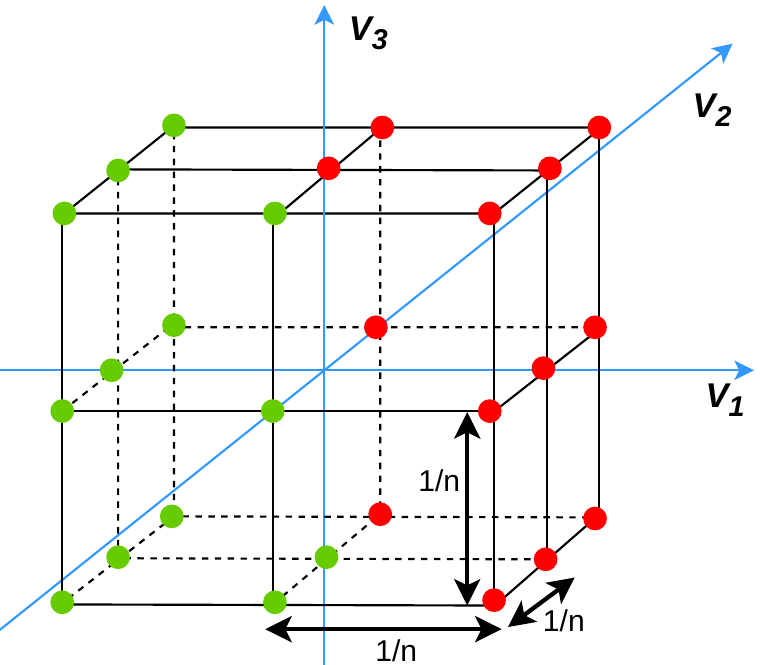} 
		\caption{Frequency points used to compute the 3D STFT. The selected frequency points are marked as red dots. The other frequency points in the green dots are ignored, as they are the complex conjugates of the selected ones.}
		\label{fig:3d-stft}
	\end{center}
\end{figure}

Given $r \in \mathbb{Z}_+$, we can define a set $\mathcal{N}_\mathbf{x}$ of 3D local neighbors of  $\mathbf{x}$ as
\begin{equation}\label{eq:1}
	\mathcal{N}_{\mathbf{x}}=\{\mathbf{y} \in\mathbb{Z}^3 \,; \|\mathbf{x}-\mathbf{y}\|_{\infty} \leq r\},
\end{equation}
where the number $n$ of neighboring feature map entries in a single dimension is given by $n=2r+1$. With this definition of neighborhood, the shape $n_t \times n_h \times n_w$ of the kernel is $n \times n \times n$.
We use the local 3D neighbors $ f(\mathbf{y}) , \forall \mathbf{y}\in \mathcal{N}_{\mathbf{x}}$ to derive the local frequency domain representation using STFT as
\begin{equation}\label{eq:2}
	F(\mathbf{v},\mathbf{x})= \sum_{\mathbf{y}\in \mathcal{N}_\mathbf{x}} f(\mathbf{y})e^{-j2\pi \mathbf{v}^T (\mathbf{x}-\mathbf{y})},
\end{equation}
where $j=\sqrt{-1}$ and $\mathbf{v} \in V$ is a 3D frequency variable. The choice of the set $V$ ($|V| = K$) can be arbitrary; we use the lowest frequency combinations with ignoring the complex conjugates as shown in Figures \ref{fig:frequency_variables} and \ref{fig:3d-stft}. Note that, due to the separability of the basis functions, 3D STFT can be efficiently computed by successively applying  1D convolutions along each dimension. 

\begin{figure}[t]
	\small
\fbox{\begin{tabular}{lll}
		$ \textbf{v}_1=[k,0,0]^T $ & $\textbf{v}_2=[k,0,k]^T$ & $\textbf{v}_3=[k,0,-k]^T$\\ 
		$ \textbf{v}_4=[0,k,0]^T$ & $\textbf{v}_5=[0,k,k]^T$ & $\textbf{v}_6=[0,k,-k]^T$\\

		$ \textbf{v}_7=[k,k,0]^T$ & $\textbf{v}_8=[k,k,k]^T$ & $\textbf{v}_9=[k,k,-k]^T$\\
		$ \textbf{v}_{10}=[k,-k,0]^T$& $\textbf{v}_{11}=[k,-k,k]^T$ & $\textbf{v}_{12}=[k,-k,-k]^T$ \\ 
		$\textbf{v}_{13}=[0,0,k]^T$ & \multicolumn{2}{l}{where $k = 1/n$}
	\end{tabular}}
	\caption{Frequency variables when $K=13$.}
	\label{fig:frequency_variables}
\end{figure}

Using the vector notation \cite{jain1989fundamentals}, we can rewrite Equation (\ref{eq:2}) to   
\begin{equation}\label{eq:3}
	F(\mathbf{v},\mathbf{x})=\mathbf{w}^T_\mathbf{v}\mathbf{f}_\mathbf{x},
\end{equation}
where $\mathbf{w}_\mathbf{v}$ is a complex valued basis function (at frequency variable $\mathbf{v}$ ) of a linear transformation, defined using ${\boldsymbol\delta}_i = \mathbf{x}-\mathbf{y}_i$ for $\mathbf{y}_i \in \mathcal{N}_\mathbf{x}$ as 
\begin{equation}\label{eq:4}
	\mathbf{w}^T_\mathbf{v}=[
	e^{-j2\pi\mathbf{v}^T{\boldsymbol\delta}_1}, \ldots, e^{-j2\pi\textbf{v}^T{\boldsymbol\delta}_{|\mathcal{N}_\mathbf{x}|}}],
\end{equation}
and $\mathbf{f}_\mathbf{x}$ is a vector containing all the elements in neighborhood $\mathcal{N}_\mathbf{x}$, defined as
\begin{equation}\label{eq:5}
	\textbf{f}_\mathbf{x}=[f(\mathbf{y}_1),\ldots, f(\mathbf{y}_{|\mathcal{N}_\mathbf{x}|})]^T.
\end{equation}
In this work, for any choice of $r$ or $n$, we consider 13 ($K=13$)
lowest non-zero frequency variables $\textbf{v}_1, \textbf{v}_2,\ldots,\textbf{v}_{13}$ as shown in Figure~\ref{fig:frequency_variables}.
Low frequency variables are used because they usually contain most information in $f$; therefore they have a better signal-to-noise ratio than high frequency components \cite{heikkila2009methods,paivarinta2011volume}. 
The local frequency domain representation for all of the above frequency variables can be aggregated as 

\begin{equation}\label{eq:6}
	\mathbf{F}_{\mathbf{x}}' = [F'(\mathbf{v}_1,\mathbf{x}), \ldots,F'(\mathbf{v}_{K},\mathbf{x})]^T .
\end{equation}

At each 3D position $\textbf{x}$, by separating the real and imaginary parts of each element, we get a vector representation of $\mathbf{F}_{\mathbf{x}}'$ as
\begin{multline}\label{eq:7}
	\mathbf{F}_{\mathbf{x}} = [\Re\{F'(\mathbf{v}_1,\mathbf{x})\}, \Im\{F'(\mathbf{v}_1,\mathbf{x})\},\\  
	\ldots,\Re\{F'(\mathbf{v}_{K},\mathbf{x})\}, \Im\{F'(\mathbf{v}_{K},\mathbf{x})\}]^T.
\end{multline}
Here, $\Re\{\cdot\}$ and $\Im\{\cdot\}$ denote the real and imaginary parts of a complex number, respectively. The corresponding $2K \times |\mathcal{N}_\mathbf{x}|$ transformation matrix can be written as
\begin{equation}\label{eq:8}
	\mathbf{W} = [\Re\{\mathbf{w}_{\mathbf{v}_1}\}, \Im\{\mathbf{w}_{\mathbf{v}_1}\}, \ldots, \Re\{\mathbf{w}_{\mathbf{v}_{K}}\}, \Im\{\mathbf{w}_{\mathbf{v}_{K}}\}].
\end{equation}
Hence, from Equations (\ref{eq:3}) and (\ref{eq:8}), the vector form of 3D STFT for all $K$ frequency points $\mathbf{v}_1,\ldots,\mathbf{v}_{K}$ can be written as
\begin{equation}\label{eq:9}
	\mathbf{F}_\mathbf{x}=\mathbf{W}^T \mathbf{f}_\mathbf{x}.
\end{equation}
Since $\mathbf{F}_\mathbf{x}$ is computed for all the elements of feature map $f$, the output feature map is with size $2K\times t\times  h\times w$ (note that we took $c = 1$). This $\mathbf{W}$ is the 3D spatio-temporal STFT kernel, and thus the depthwise STFT layer with this kernel outputs a feature map of size $2K\times t\times h \times w$ corresponding to the $K$ frequency variables. For arbitrary $c > 1$, the depthwise STFT layer extends the channel dimensions with the $3D$ output for each spatio-temporal position, which makes the output feature map of size $(c\times 2K) \times t\times h \times w$. In Figure~\ref{fig:depthwise3DSTFT}, we provide a visualization of the output of the depthwise STFT kernel for $c = 1$, $r=1$ (i.e., $n = 3$), and $K=13$.

\subsection{Variations of STFT Blocks}\label{sec:variations}


In conventional convolutional layers, each convolution kernel receives input from all channels of the previous layer, capturing spatial, temporal, and channel correlations simultaneously. If the number of input channels is large and the filter size is greater than one, it forms dense connections, leading to a high complexity. There are several micro-architectures that can be used to reduce the complexity of kernels. 

\textbf{Pointwise Bottleneck Convolutions.} \label{sec:bott_conv}
The bottleneck architecture is originally presented in \cite{he2016deep} to reduce the number of channels fed into convolutional layers, which is also used in  \cite{iandola2016squeezenet}. For spatio-temporal feature maps, $1\times 1\times 1$ (i.e., pointwise) convolutions are applied before and after a convolution with an arbitrary size kernel. These pointwise convolutions reduce and then expand the number of channels while capturing inter-channel correlations. This micro-architecture helps in reducing the number of parameters and computational cost. All micro-architectures shown in Figures~\ref{fig:a}-\ref{fig:g} use pointwise bottleneck convolutions.

\textbf{Depthwise Separable Convolutions.} \label{sec:dept_conv}
%
%
Another micro-architecture that can reduce the model complexity is depthwise separable convolutions. This is originally presented in Xception \cite{chollet2017xception}, in which a convolutional layer is divided into two parts: a convolution whose kernel only covers the spatial dimensions and a pointwise convolution whose kernel covers the channel dimension. The former is called a depthwise convolution. This can be viewed as the extreme case of group convolutions \cite{xie2017aggregated}, which divide the channel dimension into several groups. By separating the spatial and channel dimensions, connections between input and output feature maps are sparsified. Some well-known 2D CNN architectures, such as MobileNets \cite{sandler2018mobilenetv2}, and ShuffleNets \cite{zhang2018shufflenet,ma2018shufflenet}, use this micro-architecture. For 3D CNNs, depthwise convolutions cover the spatio-temporal dimensions. Figure~\ref{fig:b} illustrates the depthwise separable convolution version of Figure~\ref{fig:a}, where the following pointwise convolution is omitted.  


\textbf{Factorised 3D Convolutions.}\label{sec:fact_conv}
Particularly for 3D CNNs, yet another way to reduce the model complexity in 3D CNNs is to replace 3D (spatio-temporal) convolution kernels with separable kernels. This approach has been explored recently in a number of 3D CNN architectures proposed for video classification tasks, such as R(2+1)D \cite{tran2018closer}, S3D \cite{xie2018rethinking}, Pseudo-3D \cite{qiu2017learning}, and factorized spatio-temporal CNNs \cite{sun2015human}. The idea is to further factorize a 3D convolution kernel into ones that cover the spatial dimensions and the temporal dimension. Note that this factorization is similar in spirit to the depthwise separable convolutions as discussed above. Figures \ref{fig:a} and \ref{fig:d} show a standard 3D convolutional layer and its corresponding 3D factorization, respectively. The same idea applies to depthwise 3D convolutions as in Figures~\ref{fig:b} and \ref{fig:e}. 

\begin{figure*}[!t]
	\centering	
	\subcaptionbox{\label{fig:dw3dstftInc_st}}{\includegraphics[width=.4\columnwidth, height=0.55\columnwidth]{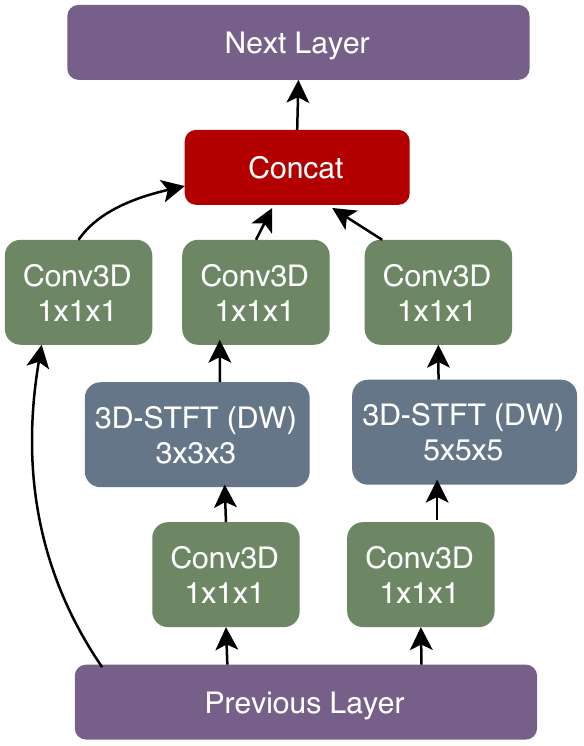}}\hspace{3em}%
	\subcaptionbox{\label{fig:dw3dstftNet_st}}{\includegraphics[width=1.1\columnwidth, height=0.55\columnwidth]{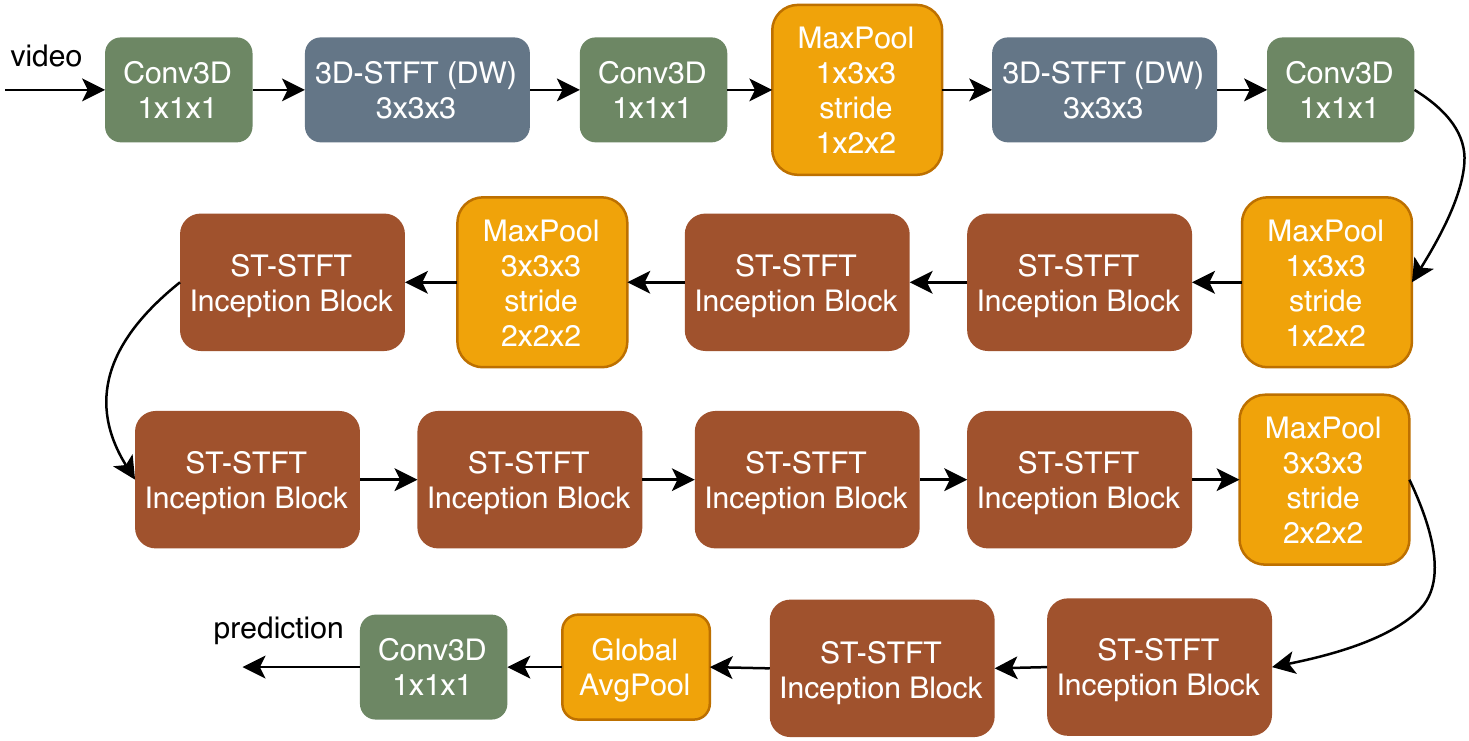}}
	\caption{(a) The ST-STFT inception block and (b) ST-STFT network architecture. }
	\label{fig:dw3dstft_st}
\end{figure*}
\begin{figure*}[!t]
	\centering	
	\subcaptionbox{\label{fig:dw3dstftInc_s}}{\includegraphics[width=.4\columnwidth, height=0.55\columnwidth]{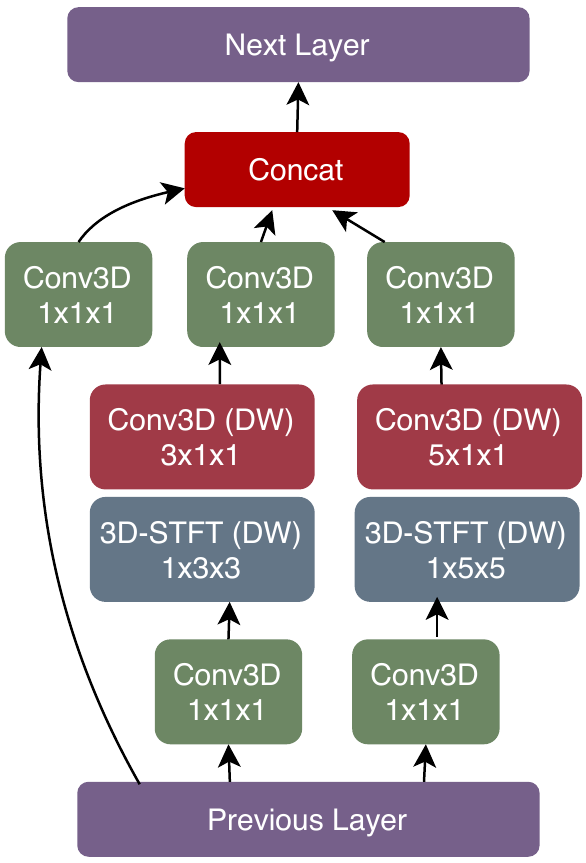}}\hspace{3em}%
	\subcaptionbox{\label{fig:dw3dstftNet_s}}{\includegraphics[width=1.3\columnwidth, height=0.6\columnwidth]{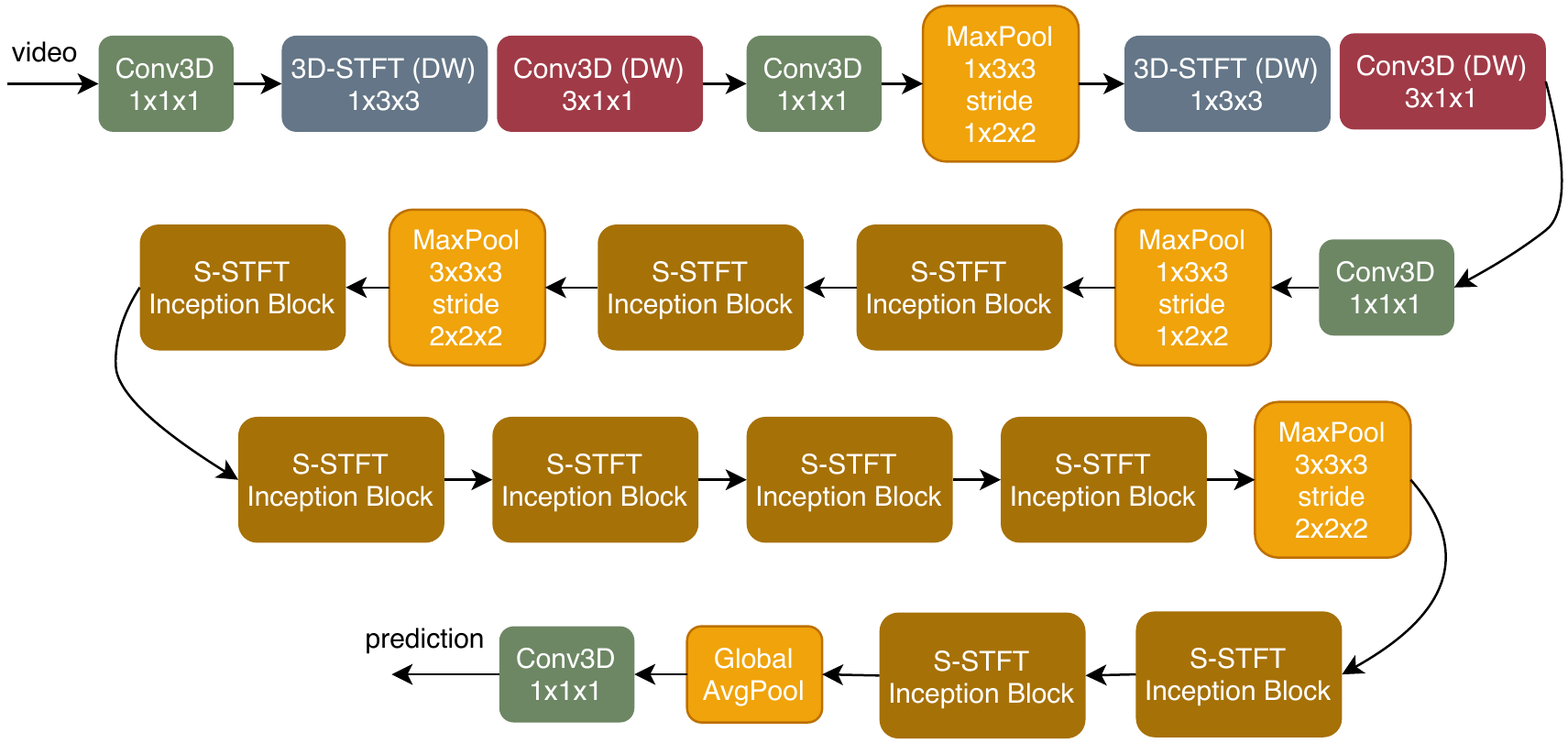}}
	\caption{(a) The S-STFT inception block and (b) S-STFT network architecture.}
	\label{fig:dw3dstft_s}
\end{figure*}
\begin{figure*}[!t]
	\centering	
	\subcaptionbox{\label{fig:dw3dstftInc_t}}{\includegraphics[width=.4\columnwidth, height=0.65\columnwidth]{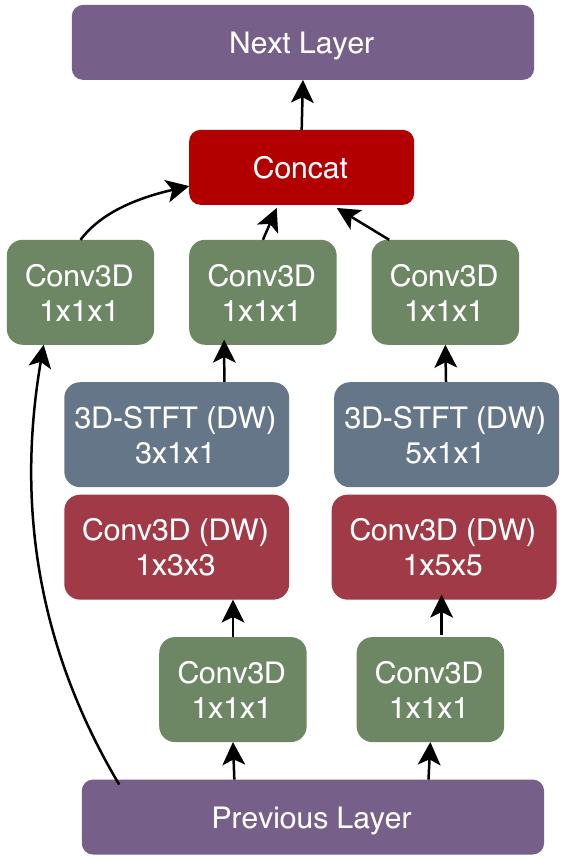}}\hspace{1em}%
	\subcaptionbox{\label{fig:dw3dstftNet_t}}{\includegraphics[width=1.3\columnwidth, height=0.6\columnwidth]{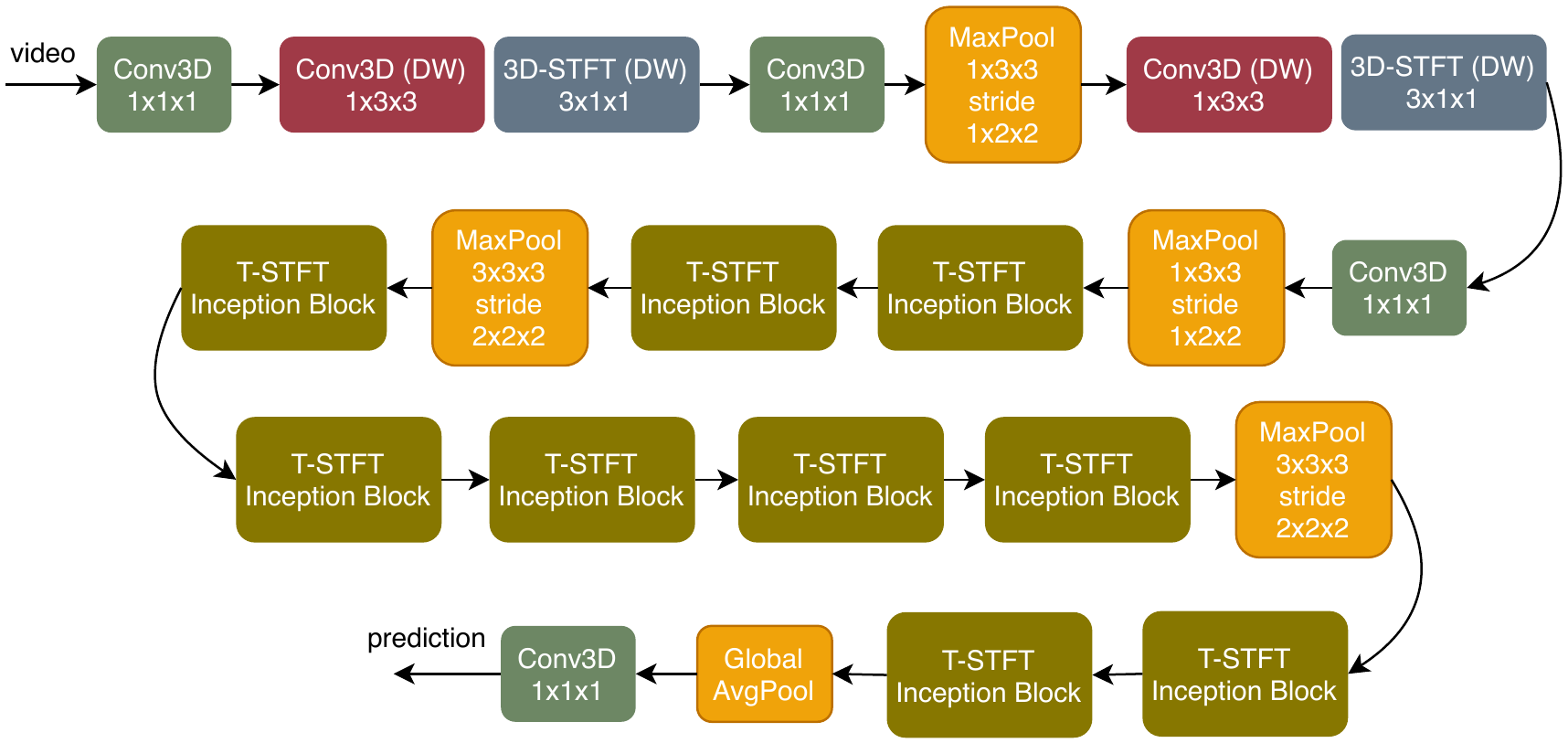}}
	\caption{(a) The T-STFT inception block and (b) T-STFT network architecture.}
	\label{fig:dw3dstft_t}
\end{figure*}


Using these three micro-architectures to reduce the model complexity, we define three different variants of the STFT blocks based on how the depthwise STFT layer is used to capture spatio-temporal correlations. 

\textbf{ST-STFT block.} The structure of this variant is based on the depthwise separable convolutional block discussed in Section~\ref{sec:dept_conv} and shown in Figure~\ref{fig:b}. The spatio-temporal information is captured using the non-trainable depthwise spatio-temporal STFT layer. Figure \ref{fig:c} illustrates the architecture. The STFT layer is sandwiched by two poinwise convolution layers, which forms a pointwise bottleneck convolution. Note that only the two pointwise convolutions are trainable. This leads to significant reduction in the number of trainable parameters and computational cost. 

\textbf{S-STFT block.} This variant is a factorization of STFT with respect to spatial and temporal dimensions as shown in Figure~\ref{fig:e}. The subscript $S$ means that the non-trainable 2D STFT kernel is used for the spatial dimensions, and a trainable 1D convolution kernel is employed for the temporal dimension. Note that the number of frequency variables are reduced accordingly. For example, ST-STFT with $K=13$ in Figure \ref{fig:frequency_variables} uses four unique frequency variables in the spatial dimensions (i.e., $\mathbf{v}_1= [k,0]^T, \mathbf{v}_2=[0,k]^T,\mathbf{v}_{3}=[k,k]^T$, and $\mathbf{v}_{4}=[k,-k]^T$) \cite{kumawat2020depthwise, heikkila2009methods}.

\textbf{T-STFT block.} The architecture of this variant is similar to the S-STFT block, but we use the non-trainable STFT kernel for the temporal dimension, and a trainable 2D convolution for the spatial dimensions as shown in Figure~\ref{fig:g}. 

\subsection{Computational Complexity}

Consider an STFT block with $c$ input and $f$ output channels, and the 3D STFT kernel of size $n\times n\times n$ with $K$ frequency variables. Below we provide theoretical analysis on the number of parameters and the computational cost of this STFT blocks. Here, $b_{s+t}$, $b_{s}$, and $b_{t}$ denote the number of bottleneck $1\times 1\times 1$ convolutions in the corresponding layers. 


\noindent\textbf{ST-STFT} 

$\# \text{parameters} = (c + 26\cdot f)\cdot b_{s+t}$

$\# \text{FLOPs} = (c+ n^3\cdot\log n^3 + 26\cdot f)\cdot h \cdot w \cdot t \cdot b_{s+t}$

\noindent\textbf{S-STFT}

$\# \text{parameters} = (c + 8n+8f)\cdot b_{s}$ 

$\#\text{FLOPs} = (c + 8n\log n+8f)\cdot h \cdot w \cdot t \cdot b_{s}$

\noindent\textbf{T-STFT}

$\# \text{parameters} = (c + n^2+2f)\cdot b_{t}$

$\#\text{FLOPs} = (c + n^2\log n+4f)\cdot h \cdot w \cdot t \cdot b_{t}$

\section{STFT Block-based Networks}\label{sec:STFTnets}

We use the BN-Inception network architecture as the backbone for designing our STFT block-based networks (referred to as X-STFT networks, where X is either ST, S, and T). We only describe the architecture of the ST-STFT network, but it is straightforward to replace them with other variants (i.e., S-STFT and T-STFT) as illustrated in Figures~\ref{fig:dw3dstft_s} and \ref{fig:dw3dstft_t}. 

The ST-STFT network consists of Inception modules called ST-STFT Inception, shown in Figure~\ref{fig:dw3dstftInc_st}. These modules are assembled upon one another with occasional max-pooling layers with stride 2 in order to reduce the resolution of feature maps. The practical intuition behind this design is that visual details should be handled at many scales and then combined hence the following layer can abstract features from various scales simultaneously. 
In our case, this architecture allows the network to choose between taking a weighted average of the feature maps in the previous layer (i.e.~by heavily weighting the $1\times1\times1$ convolutions) or focusing on local Fourier information (i.e.~by heavily weighting the depthwise STFT layer). Furthermore, each intermediate layer whether trainable or non-trainable is followed by a batch-normalization layer which is followed by an activation function. Unless otherwise mentioned, we use LeakyReLU activation with negative slope of 0.01. In Section~\ref{sec:activation}, we provide an analysis of the effect of various activation functions in the ST-STFT network and show that LeakyReLU performs the best.  The overall ST-STFT network is shown in Figure~\ref{fig:dw3dstftNet_st} with two ST-STFT blocks, followed by nine ST-STFT Inception modules (with occasional max-pooling layers) with global average pooling and softmax layers for classification.

\section{Experimental Setting}\label{sec:experiments}

\subsection{Datasets}\label{sec:datasets}
\begin{figure}[t]
	\begin{center}
		\includegraphics[width=\columnwidth, height=0.3\columnwidth]{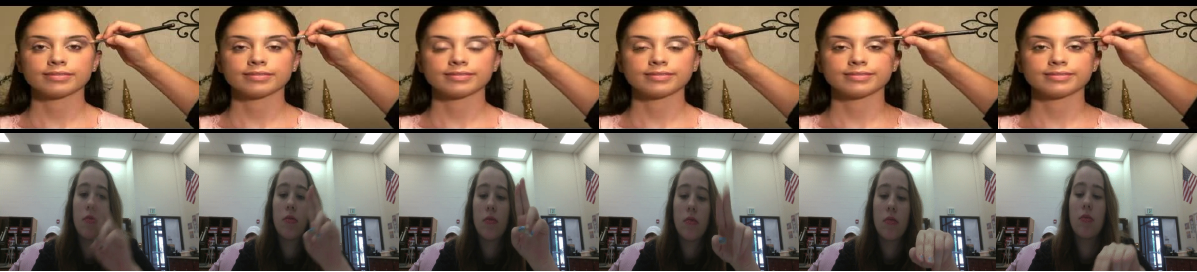} 
		\caption{Difference between scene-related vs temporal-related datasets. Top: \texttt{Apply Eye Makeup} action class from the UCF-101 dataset. Only one frame is enough to predict the label.  Bottom: \texttt{Sliding Two Fingers Down} action class from the Jester dataset. Reversing the order of frames gives the opposite label which is \texttt{Sliding Two Fingers Up}.}
		\label{fig:scene_temp}
	\end{center}
\end{figure}

We evaluate the performance of our proposed X-STFT networks on seven publicly available action recognition datasets. Following \cite{jiang2019stm}, we group these datasets into two categories:  scene-related datasets and temporal-related datasets. The former includes the datasets, in which the spatial cues, such scenes, objects, and background, are dominant for action recognition. In many samples, the action can be correctly recognized even from a single frame. For this category, the temporal signal is not much informative. Kinetics-400 \cite{carreira2017quo}, UCF-101 \cite{soomro2012ucf101}, and HMDB-51 \cite{kuehne2011hmdb} fall into this category. The latter requires to use temporal interactions among objects for recognizing actions. Thus, the temporal information, as well as the spatial information, plays an important role. This category includes Jester \cite{yogajournal}, Something$^2$ v1 \& v2 \cite{goyal2017something}, and Diving-48 \cite{li2018resound}. Figure \ref{fig:scene_temp} shows the difference between them. We essentially focus on the temporal-related datasets since the proposed method is designed for effectively encoding spatio-temporal information. Nonetheless, X-STFT networks achieve competitive results even over the scene-related datasets.

\begin{table}[!t]
	\centering
		\caption{Details of the benchmark human action recognition datasets used for evaluation. The top four, i.e. Jester, Something$^2$ v1, Something$^2$ v2, and Diving-48 are temporal-related datasets. The remaining three, i.e. Kinetics-400, UCF-101, and HMDB-51 are scene-related datasets.  }
	\begin{tabular}{lcccc}  
		\toprule[0.15em]
		\textbf{Model}   & \textbf{\# Classes} & \textbf{\# Train}  & \textbf{\# Val} & \textbf{\# Test} \\
		\midrule[0.15em]
		Jester~\cite{yogajournal} & 27 &  1,18,562 & 14,787 & 14,743\\
		Something$^2$ v1~\cite{goyal2017something} & 174 &  86,017 & 11,522 & 10,960\\
		Something$^2$ v2~\cite{goyal2017something} &  174 &  1,68,913 & 24,777 & 27,157 \\
		Diving-48~\cite{li2018resound} &  48 &  16,067 &  2,337 & - \\
		Kinetics-400~\cite{carreira2017quo} &  400 & 2,46,535  & 19,907 & 38,686 \\
		UCF-101~\cite{soomro2012ucf101} &  101 &  9,537 & 3,783 & - \\
		HMDB-51~\cite{kuehne2011hmdb} &  51 &  3,570 & 1,530 & - \\
		\bottomrule 
	\end{tabular}
	\label{tab:datasets}
\end{table}

\subsection{Implementation Details}\label{sec:train_details}

For fair comparison, we broadly follow the configuration in \cite{jiang2019stm}. We detail it to make the paper self-contained.

\noindent\textbf{Input and augmentation:} Let $T$ be the number of frames per input video. We first uniformly divide the input video into $T$ segments. Next, we randomly sample one frame per segment in order to form an input sequence with $T$ frames. If a video contains less than $T$ frames, we apply loop padding. For spatial data augmentation, we randomly choose for each input sequence a spatial position and a scale in order to perform multi-scale cropping as in \cite{hara2018can}, where the scale is picked out from the set $\{1, \frac{1}{2^{1/4}}, \frac{1}{2^{3/4}}, \frac{1}{2}\}$. For Kinetics-400, UCF-101, HMDB-51, and Diving-48 datasets, we horizontally flip all frames in each input sequence with 50\% probability. Note that this random horizontal flipping is not applied to the other datasets as few of the classes form a symmetrical pair. For example, in Jester, \texttt{Swiping Left} and \texttt{Swiping Right} are a symmetrical pair. Similarly, \texttt{Sliding Two Fingers Left} and \texttt{Sliding Two Fingers Right} are another symmetrical pair. An input sequence is consequently packed into a tensor in $\mathbb{R}^{3 \times T \times 112 \times 112}$.

\noindent\textbf{Training:} We use stochastic gradient descent (SGD) as optimizer, categorical cross-entropy as loss function, and the mini-batch size of 32. The hyperparameters for momentum, dampening, and weight decay are set to 0.9, 0.9, and $10^{-3}$, respectively. All the trainable weights are initialized with orthogonal initializer. When training from scratch, we update networks for 160 epochs, starting with learning rate of 0.1 and decreasing it by a factor of 10 after every 40 epochs. When training from pre-trained weights, we update for 60 epochs, starting with learning rate of 0.01 and decreasing it by a factor 10 after every 20 epochs. 

\noindent\textbf{Inference:} Following \cite{xie2018rethinking}, we sample from each input video equi-distant $T$ frames without random shift, producing an input sequence. We then crop each frame in the sequence with the square region at the center of the frame without scaling. Finally, the sequence is rescaled and packed into $3\times T \times 112 \times 112$. Probability class score for every clip is calculated using trained models.  For all the datasets, we report accuracy on the validation split and perform corresponding comparisons with other works.

\begin{table}[!htb]
	
		\centering
		\caption{Comparison with similar backbone networks. All networks use the same BN-Inception network architecture as backbone. Fact.+DW denotes factorized+DW block (Figure~\ref{fig:e}).}
		\begin{tabular}{lccccc}  
			\toprule
			& & &  &\multicolumn{2}{c}{\textbf{Something$^2$v1}} \\
			\cmidrule(r){5-6}
			\textbf{Model} & \textbf{Frames} & \textbf{Params} & \textbf{FLOPs}  & \multicolumn{1}{c}{\textbf{Top 1}} & \multicolumn{1}{c}{\textbf{Top 5}} \\
			\midrule
			I3D \cite{carreira2017quo} &16 & 12.90M & 26.97G & 34.6 & 64.7 \\
			ReLPV \cite{kumawat2019lp}  & 16 & 4.78M & - & 38.64 & 69.17 \\
			Fact. + DW  & 16 & 6.28M & 10.30G & 42.57 & 72.68 \\
			ST-STFT & 16 & 5.84M & 10.63G & 43.78
			 & 73.59\\
			S-STFT & 16 & 6.03M & 10.39G & 42.91 & 72.67\\
			T-STFT  & 16 & 6.27M & 10.30G & \textbf{46.74} & \textbf{76.15}\\
			\midrule
			I3D \cite{carreira2017quo} & 32 & 12.90M & 53.95G & 43.6 & 73.8 \\
			ReLPV \cite{kumawat2019lp} & 32 & 4.78M & - & 40.69 & 71.20 \\
			Fact. + DW & 32 & 6.28M & 20.79G & 44.62 & 74.37 \\
			ST-STFT   & 32 & 5.84M & 21.26G & 46.15 & 75.99\\
			S-STFT  & 32 & 6.03M & 20.79G  & 45.10 & 74.86\\
			T-STFT  & 32 & 6.27M & 20.60G & \textbf{48.54} & \textbf{78.12}\\
			\midrule
			I2D & 64 & 6.79M & - & 34.4 & 69.0 \\
			I3D \cite{carreira2017quo} & 64 & 12.90M & 107.89G & 45.8 & 76.5 \\
			S3D \cite{xie2018rethinking} & 64 & 8.77M  & 66.38G & 47.3 & 78.1 \\
			S3D-G \cite{xie2018rethinking} & 64 & 11.56M  & 71.38G & 48.2 & 78.7 \\
			ReLPV \cite{kumawat2019lp} & 64 & 4.78M & - & 43.01 & 73.32 \\
			Fact. + DW & 64 & 6.28M & 41.21G & 46.75 & 76.52 \\
			ST-STFT  & 64 & 5.84M & 42.52G & 48.23 & 77.79\\
			S-STFT  & 64 & 6.03M & 41.59G & 47.12 & 76.91\\
			T-STFT  & 64 & 6.27M & 41.21G & \textbf{50.63} & \textbf{79.59}\\
			\bottomrule
		\end{tabular}
		\label{tab:backbone}
	\end{table}

\section{Ablation Study}\label{sec:ablation_studies}
In this section, we present comparative performance studies of the STFT  blocks-based networks and their variant on the Something$^2$ v1 action recognition dataset. We will utilize the network architectures explained in Section~\ref{sec:STFTnets}. Furthermore, all the networks are trained from scratch using the training and inference methodologies in Section~\ref{sec:train_details}. 

\subsection{Comparison with similar backbone networks}\label{sec:backbone}
As mentioned earlier in Section~\ref{sec:STFTnets}, we use the BN-Inception architecture as backbone for developing our X-STFT networks. Various state-of-the-art networks use BN-Inception as backbone, including I3D \cite{carreira2017quo}, S3D \cite{xie2018rethinking}, and S3D-G \cite{xie2018rethinking}. Table \ref{tab:backbone} compares their performances with our X-STFT networks on the Something$^2$v1 dataset. We observe that, compared to I3D that uses conventional 3D convolutional layers, X-STFT's use roughly 2 times less parameters and 2.5 times less computations. Similarly, when compared with S3D and S3D-G that use factorized 3D convolutional kernels, our networks use 1.4 and 1.8 times less parameters, respectively. Furthermore, X-STFT's use 1.5 and 1.6 times less computations when compared to S3D and S3D-G, respectively. Apart from the computation and parameter savings, the X-STFT networks, especially T-STFT, consistently achieve higher accuracy levels. As mentioned earlier in Section~\ref{sec:introduction}, the ST-STFT block is an extension of our ReLPV block \cite{kumawat2019lp} published in CVPR 2019. Unlike ST-STFT block, the ReLPV block use extreme bottleneck pointwise convolutions and the STFT kernel is not applied in depthwise fashion. For fair comparison, we replace the ST-STFT blocks with the ReLPV blocks in the ST-STFT network. We observe that applying STFT in depthwise fashion improves the accuracy of the ST-STFT network  by almost 5\% when compared with the one containing ReLPV blocks. Furthermore, for strict baseline comparisons, we use the best performing T-STFT network and replace its fixed STFT kernel with trainable standard depthwise convolution layer and call this baseline network as Fact.+DW. We observe that using fixed STFT kernel for capturing temporal information improves the accuracy by almost 4\%, compared with the case when trainable weights were used.

An important observation from these results is that the Fourier coefficients are efficient in encoding the motion representations. This is evident from the fact that the T-STFT and ST-STFT networks always achieve better accuracy for different numbers of frames when compared to the S-STFT network that uses trainable depthwise convolutions for capturing temporal information.     

	\begin{table}[t]
			\centering
	\caption{Comparison with state-of-the-art depthwise 3D convolution-based networks. Number of frames used for evaluation is 32. For inference only single center crop is used.}
	\begin{tabular}{lcccc}  
		\toprule[0.15em]
		& &  &\multicolumn{2}{c}{\textbf{Something$^2$v1}} \\
		\cmidrule(r){4-5}
		\textbf{Model}   & \textbf{Params} & \textbf{FLOPs}  & \multicolumn{1}{c}{\textbf{Top 1}} & \multicolumn{1}{c}{\textbf{Top 5}} \\
		\midrule[0.15em]
		3D ShuffleNetV1 1.5x \cite{kopuklu2019resource} & 2.31M & 0.48G  & 32.12 & 61.43 \\
		3D ShuffleNetV2 1.5x  \cite{kopuklu2019resource}  & 2.72M & 0.44G  & 31.09  & 60.30 \\
		3D MobileNetV1 1.5x  \cite{kopuklu2019resource} & 7.56M & 0.54G  & 27.88 & 55.77 \\
		3D MobileNetV2 0.7x  \cite{kopuklu2019resource} & 1.51M & 0.51G  & 27.51 & 55.31\\
		\midrule
		3D ShuffleNetV1 2.0x  \cite{kopuklu2019resource}  & 3.94M & 0.78G  & 33.91 & 62.52 \\
		3D ShuffleNetV2 2.0x  \cite{kopuklu2019resource} & 5.76M & 0.72G  & 31.89 & 61.02 \\
		3D MobileNetV1 2.0x  \cite{kopuklu2019resource}  & 13.23M & 0.92G  & 29.75 & 56.89\\
		3D MobileNetV2 1.0x  \cite{kopuklu2019resource} & 2.58M & 0.91G & 30.78 & 59.76\\
		\midrule
		ir-CSN-101 \cite{tran2019video} & 22.1M & 56.5G  & 47.16 & -\\
		ir-CSN-152 \cite{tran2019video} & 29.6M & 74.0G & 48.22 & -\\
		\midrule 
		ST-STFT   & 5.84M & 21.26G & 46.15 & 75.99\\
		S-STFT   & 6.03M & 20.79G & 45.10 & 74.86\\
		T-STFT   & 6.27M & 20.60G & \textbf{48.54} & \textbf{78.12} \\
		\bottomrule
	\end{tabular}
	\label{tab:depthwise}
\end{table}

\subsection{Comparison with state-of-the-art depthwise 3D convolution-based networks}
In this study, we compare our X-STFT networks with some of the state-of-the-art 3D CNNs that use depthwise 3D convolutions for capturing the spatio-temporal interactions. Table~\ref{tab:depthwise} compares the performance of such networks with our proposed networks on the Something$^2$ v1 dataset. For fair comparison, all networks use 32 frames for training and testing. We observe that the direct 3D extension \cite{kopuklu2019resource} of depthwise convolution-based 2D CNN architectures such as MobileNet \cite{howard2017mobilenets} and ShuffleNet \cite{zhang2017shufflenet,zhang2018shufflenet} use comparable parameters and significantly less computations when compared to the X-STFT networks. However, they achieve poor accuracy levels. Other models that are based on depthwise convolutions, such as ir-CSN \cite{tran2019video}, are specially developed for action recognition tasks and are able to achieve comparable accuracy levels to our networks. However, they need to be deep and thus require a large number of parameters and computations. For example, ir-CSN-152 achieves a comparable accuracy compared to the T-STFT network, although it uses 4.7 times more parameters and 3.5 times more computations than the T-STFT network.

\begin{table}[htbp]
 \centering
		\caption{Effect of different activation functions on the performance of the ST-STFT network. Number of frames used for evaluation is 16.}
		\begin{tabular}{lccc}  
			\toprule
			& &\multicolumn{2}{c}{\textbf{Something$^2$v1}} \\
			\cmidrule(r){3-4}
			\textbf{Model}   & \textbf{Activation}  & \multicolumn{1}{c}{\textbf{Top 1}} & \multicolumn{1}{c}{\textbf{Top 5}} \\
			\midrule
			ST-STFT  & ReLU  &  42.62 &	72.89\\
			ST-STFT  & LeakyReLU  &  \textbf{43.78} & \textbf{73.59}\\
			ST-STFT  &  SELU &  32.83 & 61.83 \\
			ST-STFT  &  ELU &  32.39 & 61.87\\
			\bottomrule
		\end{tabular}
		\label{tab:activation}
		\vspace{.2in}
\end{table}

\begin{table*}[!t]
	\centering
	\caption{Performance of the X-STFT networks on the Something$^2$v1 and v2 datasets compared with the state-of-the-art methods. FLOPs values are in the format- FLOPs per clip $\times$ \# of crops per clip $\times$ \# of clips sampled from each video.}
	\begin{tabular}{lcccccccccc}  
		\toprule[0.15em]
		& & & & & & & \multicolumn{2}{c}{\textbf{Something$^2$v1}}  &\multicolumn{2}{c}{\textbf{Something$^2$v2}} \\
		\cmidrule(r){8-9}
		\cmidrule(r){10-11}
		\multicolumn{1}{c}{\textbf{Model}} & \multicolumn{1}{c}{\textbf{Backbone}} &  \multicolumn{1}{c}{\textbf{Input}} & \multicolumn{1}{c}{\textbf{Pre-training}} & \multicolumn{1}{c}{\textbf{Params}} &
		\multicolumn{1}{c}{\textbf{Frames}} &
		 \multicolumn{1}{c}{\textbf{FLOPs}} & \multicolumn{1}{c}{\textbf{Top 1}} & \multicolumn{1}{c}{\textbf{Top 5}} & \multicolumn{1}{c}{\textbf{Top 1}} & \multicolumn{1}{c}{\textbf{Top 5}} \\
		\midrule
		\midrule
		I3D \cite{carreira2017quo} & 3D ResNet-50 & RGB & ImageNet & - & 32 &153G$\times$3$\times$2 & 41.6 & 72.2 & - &-\\
		NL-I3D \cite{carreira2017quo} & 3D ResNet-50 & RGB & + & - & 32 &168G$\times$3$\times$2 & 44.4 & 76.0 & - &-\\
		I3D+GCN \cite{carreira2017quo} & 3D ResNet-50 & RGB & Kinetics & - & 32 & 303G$\times$3$\times$2 &46.1 & 76.8 & - &-\\	
		\midrule
		S3D-G  \cite{xie2018rethinking} & BN-Inception & RGB & ImageNet & 11.56M & 64 & 71.38G$\times$1$\times$1 &48.2 & 78.7 & - & -\\
		\midrule
		ECO \cite{zolfaghari2018eco} & BN-Inception & RGB  & Kinetics & 47.5M& 8 & 32G$\times$1$\times$1 & 39.6 & - & - & -\\ 	
		ECO \cite{zolfaghari2018eco} & + & RGB  & Kinetics & 47.5M& 16 & 64G$\times$1$\times$1 & 41.4 & - & - & -\\ 
		ECO$_{En}$ \cite{zolfaghari2018eco} & 3D ResNet-18  & RGB & Kinetics & 150M & 92 & 267G$\times$1$\times$1 &46.4 & - & - & -\\ 
		ECO$_{En}$ Two stream \cite{zolfaghari2018eco} &   & RGB+Flow & Kinetics & 300M & 92+92 & - & 49.5 & - & - & -\\
		\midrule
		ir-CSN-101 \cite{tran2019video} & 3D ResNet-50 & RGB & \xmark & 22.1M & 8 & 56.5G$\times$1$\times$10 & 48.4 & - & - & -\\
		ir-CSN-152 \cite{tran2019video} & 3D ResNet-50 & RGB & \xmark & 29.6M & 16 & 74G$\times$1$\times$10 & 49.3 & - & - & -\\ 
		\midrule
		\midrule
		TSN \cite{wang2016temporal} & ResNet-50 & RGB & Kinetics & 24.3M & 8 & 16G$\times$1$\times$1 &19.5 & 46.6 & 27.8 & 57.6\\
		TSN \cite{wang2016temporal} & ResNet-50 & RGB & Kinetics & 24.3M  & 16 & 33G$\times$1$\times$1 &19.7 & 47.3 & 30.0 & 60.5\\
		\midrule
		TRN Multiscale \cite{zhou2018temporal} & BN-Inception & RGB & ImageNet & 18.3M & 8 & 16.37G$\times$1$\times$1 &34.4 & - & 48.8 & 77.64\\
		TRN Two stream \cite{zhou2018temporal} & BN-Inception & RGB+Flow & ImageNet &  36.6M & 8+8 & - & 42.0 & - & 55.5 & 83.1\\
		\midrule
		TSM \cite{lin2019tsm} & ResNet-50 & RGB & Kinetics & 24.3M & 8 & 33G$\times$1$\times$1 & 45.6 & 74.2 & 59.1 & 85.6\\
		TSM \cite{lin2019tsm} & ResNet-50 & RGB & Kinetics & 24.3M & 16 & 65G$\times$1$\times$1 & 47.2 & 77.1 & 63.4 & 88.5\\
		\midrule
		MFNet-C101 \cite{lee2018motion} & ResNet-101 & RGB & \xmark & 44.7M & 10 & - &43.9 & 73.1 & -& -\\ 
		\midrule
		STM \cite{jiang2019stm} & ResNet-50 & RGB & ImageNet & 26M & 8 & 33G$\times$3$\times$10 & 49.2 & 79.3 & 62.3 & 88.8\\
		STM \cite{jiang2019stm} & ResNet-50 & RGB & ImageNet & 26M & 16 & 67G$\times$3$\times$10 & 50.7 & 80.4 & 64.2 & 89.8\\
		\midrule
		GST \cite{luo2019grouped} & ResNet-50 & RGB & ImageNet & 29.6M & 8 & 29.5G$\times$1$\times$1 & 47.0 & 76.1 & 61.6 & 87.2\\
		GST \cite{luo2019grouped} & ResNet-50 & RGB & ImageNet & 29.6M & 16 & 59G$\times$1$\times$1 & 48.6 & 77.9 & 62.6 & 87.9\\
		\midrule
		\midrule
		ST-STFT  & BN-Inception & RGB & \xmark & 5.84M & 64 & 42.52G$\times$1$\times$1 & 48.23 & 77.79 & 61.56 & 87.81\\
		S-STFT  & BN-Inception & RGB & \xmark & 6.03M & 64 & 41.59G$\times$1$\times$1 & 47.12 & 76.91 & 60.34 & 86.56 \\
		T-STFT  & BN-Inception & RGB & \xmark & 6.27M & 64 & 41.21G$\times$1$\times$1 & 50.63 & 79.59 & 63.05 & 89.31\\
		\midrule
		T-STFT  & BN-Inception & RGB & Kinetics & 6.27M & 16 & 10.30G$\times$1$\times$1 & 48.22 & 77.89 & 61.72 & 87.56\\		
		T-STFT  & BN-Inception & RGB & Kinetics & 6.27M & 32 & 20.60G$\times$1$\times$1 & 50.25 & 80.23 & 63.22 & 89.34\\		
		T-STFT  & BN-Inception & RGB & Kinetics & 6.27M & 64 & 41.21G$\times$1$\times$1 & \textbf{52.42} & \textbf{81.8} & \textbf{64.66} & \textbf{90.84}\\		
		\bottomrule[0.15em]
	\end{tabular}
	\label{tab:ssv1v2}
\end{table*}
\vspace{-2em}
\subsection{Choice of activation functions}\label{sec:activation}
An important hyperparameter in deep CNNs is the activation function. In this study, we explore various activation functions that suite for the X-STFT networks. We assume that the results of this study can be agnostic to the variations of the networks; therefore, we only evaluate the performance on the ST-STFT network. Table~\ref{tab:activation} compares the performances on the Something$^2$ v1 dataset when a single type of activation function is used after every convolution layer. For fair comparison, all networks use 16 frames for training and testing. We observe that the LeakyReLU activation with a small negative slope of 0.01 achieves the best performance, which is followed by the ReLU activation.

\section{Results}\label{sec:results_AR}

\subsection{Results on Temporal-Related Datasets}\label{sec:results_ART}
We first compare our X-STFT networks with the existing state-of-the-art methods on the Something$^2$ v1 and v2 action recognition datasets. Table~\ref{tab:ssv1v2} provides a comprehensive comparison of these methods in terms of various statistics such as the input type, parameters, inference protocols, FLOPs, and classification accuracy. Following \cite{jiang2019stm}, we separate existing methods into two groups. The first group contains methods that are based on 3D CNNs and its factorized variants, including I3D \cite{carreira2017quo}, S3D-G \cite{xie2018rethinking}, ECO \cite{zolfaghari2018eco}, and ir-CSN \cite{tran2019video}.  The second group contains methods that are based on 2D CNNs with various temporal aggregation modules, including TSN \cite{wang2016temporal}, TRN \cite{zhou2018temporal}, TSM \cite{lin2019tsm}, MFNet \cite{lee2018motion}, STM \cite{jiang2019stm}, and GST \cite{luo2019grouped}. Among these groups, the current state-of-the-art is the STM model, which achieves the top-1 accuracy of 50.7\% and 64.2\% on Something$^2$ v1 and v2 datasets, respectively. We observe that our proposed STFT network outperforms the STM network by a margin of 1.72\% and 0.46\% on the  Something$^2$ v1 and v2 datasets, respectively.
Furthermore, despite the fact that the STFT network use 64 frame input when compared to 16 frames in STM, the STFT network uses 4.1 times less parameters and 1.6 times less computations. In the first group, the most efficient models in terms of parameters and computations (with competitive accuracy) are S3D-G and ir-CSN-152. Compared to the S3D-G (64 frames) and ir-CSN-152 (16 frames) models, our T-STFT (64 frames) model uses 1.8 and 4.7 times less parameters, respectively. Furthermore, they it uses 1.7 and 1.8 times less computations when compared to the S3D-G and ir-CSN-152 models, respectively. Similarly, in the second group, the most efficient models with competitive accuracy are STM, GST, and TSM. Compared to the GST and TSM networks, the T-STFT model uses 4.7 and 3.8 times less parameters, respectively. Furthermore, it uses 1.4 and 1.5 times less computations when compared to the GST and TSM networks, respectively. 

\begin{table*}[!t]
	\begin{minipage}{.55\linewidth}
	\centering
	\caption{Performance of the X-STFT networks on the Diving-48 dataset compared with the state-of-the-art methods.}
	\begin{tabular}{lcccc}  
		\toprule
		\textbf{Model}   & \textbf{Input}  & \textbf{Pre-training} & \textbf{Frames} & \textbf{Diving-48} \\
		\midrule
		
		TSN \cite{wang2016temporal}		    & RGB 		& ImageNet 		& 16	 	& 16.8\\
		
		TSN \cite{wang2016temporal} 			& RGB+Flow  & ImageNet  		& 16	& 20.3 \\
		TRN [\cite{zhou2018temporal} 			& RGB+Flow  & ImageNet 		& 16	    & 22.8\\
		C3D \cite{li2018resound,tran2015learning}	    &  RGB 		& Sports1M 		&	64	& 27.6\\
		R(2+1)D \cite{tran2018closer,bertasius2018learning} 	& RGB 		&  Kinetics 		&	64	& 28.9\\
		Kanojia	\emph{et al} \cite{kanojia2019attentive}			& RGB 		& ImageNet  	&	64	& 35.6 \\	
		\midrule
		Bertasius \emph{et al} \cite{bertasius2018learning} & Pose		& PoseTrack  		& -	& 17.2 \\
		Bertasius \emph{et al} \cite{bertasius2018learning} & Pose+Flow & PoseTrack 		&-	& 18.8 \\
		Bertasius \emph{et al} \cite{bertasius2018learning} & Pose+DIMOFS & PoseTrack  		&-	& 24.1 \\
		\midrule
		P3D-ResNet18 \cite{qiu2017learning} & RGB & ImageNet & 16 & 30.8  \\
		C3D-ResNet18 & RGB & ImageNet & 16 & 33.0  \\
		GST-ResNet18 \cite{luo2019grouped} & RGB & ImageNet & 16 & 34.2  \\
		\midrule
		P3D-ResNet50 \cite{qiu2017learning} & RGB & ImageNet & 16 & 32.4 \\
		C3D-ResNet50 & RGB & ImageNet & 16 & 34.5 \\
		GST-ResNet50 \cite{luo2019grouped} & RGB & ImageNet & 16& 38.8 \\
		\midrule
		ST-STFT  & RGB  & Kinetics & 64 &	39.4 \\
		S-STFT  & RGB   &Kinetics & 64 & 36.1\\
		T-STFT  & RGB   &Kinetics & 64 & \textbf{44.3} \\
		\bottomrule
	\end{tabular}
	\label{tab:diving48}
\end{minipage}%
	\begin{minipage}{.45\linewidth}
		\centering
		\footnotesize
		\caption{Performance of the X-STFT networks on the Jester dataset compared with the state-of-the-art methods.}
		\begin{tabular}{lccc}  
			\toprule
			\textbf{Model}  & \textbf{Pre-training}  & \textbf{Frames}   &\textbf{\textbf{Jester}} \\
			\midrule
			TSN \cite{wang2016temporal} & ImageNet & 16  & 82.30  \\
			TSM \cite{lin2019tsm} & ImageNet  & 16  & 95.30  \\		
			TRN-Multiscale \cite{zhou2018temporal}  & ImageNet & 8  & 95.31  \\
			MFNet-C50 \cite{lee2018motion}  & \xmark & 10  &  96.56  \\
			MFNet-C101 \cite{lee2018motion}  & \xmark & 10 &   96.68  \\
			STM \cite{jiang2019stm}  & ImageNet & 16  & 96.70  \\	
			\midrule
			3D-SqueezeNet \cite{kopuklu2019resource}  & \xmark  &  16 & 90.77 \\
			3D-MobileNetV1 2.0x \cite{kopuklu2019resource}  & \xmark  & 16  &   92.56 \\
			3D-ShuffleNetV1 2.0x \cite{kopuklu2019resource}  & \xmark &  16  &   93.54  \\
			3D-ShuffleNetV2 2.0x \cite{kopuklu2019resource}  & \xmark &  16  &   93.71\\
			3D-MobileNetV2 1.0x \cite{kopuklu2019resource}  & \xmark &  16  &   94.59 \\	
			\midrule
			3D ResNet-18 \cite{kopuklu2019resource}  & \xmark & 16   & 93.30 \\
			3D ResNet-50 \cite{kopuklu2019resource}  & \xmark &   16  & 93.70  \\
			3D ResNet-101 \cite{kopuklu2019resource}  & \xmark &  16  & 94.10 \\
			3D ResNeXt-101 \cite{kopuklu2019resource}  & \xmark & 16  & 94.89 \\
			\midrule
			ST-STFT & \xmark &  16 & 96.51 \\
			S-STFT   & \xmark &  16 & 96.19 \\
			T-STFT  & \xmark &  16 & 96.85 \\
			T-STFT  & Kinetics & 16 & \textbf{96.94} \\
			\bottomrule
		\end{tabular}
		\label{tab:jester}
	\end{minipage}
\end{table*}

In Table~\ref{tab:diving48}, we present quantitative results on the Diving-48 dataset. Note that the Diving-48 dataset is a fine-grained action recognition dataset where all videos correspond to a single activity, i.e. diving. It contains 48 forms (or classes) of diving which the model should learn to discriminate. Such property makes the task of recognizing actions more challenging. We notice that our T-STFT model surpasses all the methods by a significant margin in terms of accuracy, number of parameters, and computations. For example, it outperforms the current state-of-the-art GST-ResNet50 (38.8\%) network by 5.5\%. Furthermore, despite using a 64 frame input it uses 1.4 times less computations and 4.7 times less parameters when compared to the GST-ResNet50 network which uses 16 frames.

In Table~\ref{tab:jester}, we present quantitative results on the Jester dataset which consists of videos of people performing different hand gestures. Note that all the methods on this dataset use RGB frames only. Since Jester does not have enough frames per video, we perform our proposed method with only 16 frames in this case. We observe that, in terms of accuracy, our T-STFT model outperforms all its existing counterparts. Furthermore, compared to the current state-of-the-art STM \cite{jiang2019stm} model, it uses 1.6 times less computations and 4.1 times less parameters.

\subsection{Results on Scene-Related Datasets}
In this section, we compare the performance of our STFT-based networks with existing state-of-the-art models on some scene-related datasets. Note that we will skip the space-time complexity comparisons here as they have been already discussed in Section~\ref{sec:results_ART}, where we showed that the STFT-based models significantly reduce parameters and computations. Also, we have used 64 frames in all scene-related dataset evaluation.

In Table~\ref{tab:kinetics}, we provide quantitative results on Kinetics-400 dataset. We observe that our best performing T-STFT network achieves the accuracy of 61.1\%. We believe that this relatively lower accuracy of the STFT-based models on the Kinetics dataset can be attributed to the following factors.
\begin{enumerate*}[label=(\arabic*), ref=\arabic*]
\item Unlike existing state-of-the-art models that use pre-trained ImageNet weights in order to achieve a higher accuracy, we train our STFT-based models from scratch.
\item As observed earlier in Section~\ref{sec:results_ART}, the STFT-based models use very few trainable parameters (3 to 5 times less) when compared to other state-of-the-art models. Furthermore, the Kinetics dataset is a very large dataset with 400 classes and 0.24M training examples. We reason that these conditions led to the underfitting of our models on the Kinetics dataset. Such behavior is common among networks when trained on large datasets. For example, the R(2+1)D network with ResNet-18 backbone performs worst than the version with ResNet-34 backbone. Our T-STFT network outperforms the R(2+1)D (backone ResNet-18) network while using 5.2 times less parameters. 
\end{enumerate*}

In Table~\ref{tab:ucf-hmdb}, we present results of our generalization study of our Kinetics pre-trained models on the the UCF-101 and HMDB-51 datasets. For both UCF-101 and HMDB-51, we evaluate our
models over three splits and report the averaged accuracies. From Table~\ref{tab:ucf-hmdb}, we observe that, compared with the ImageNet or Kinetics pre-trained
models, ImageNet+Kinetics pre-trained can notably enhance the performance on small datasets. Also, our Kinetics pre-tranied STFT-based models outperform the models pre-trained on the Sports-1M, Kinetics, or ImageNet datasets. Furthermore, they achieve comparable accuracies when compared with the ImageNet+Kinetics pre-trained models.

\begin{table}[!t]
	\centering
	\caption{Performance of the X-STFT networks on the Kinetics-400 dataset compared with the state-of-the-art methods.}
	\begin{tabular}{lcccc}  
		\toprule
		& & &\multicolumn{2}{c}{\textbf{Kinetics-400}} \\
		\cmidrule(r){4-5}
		\textbf{Model} & \textbf{Backbone}  & \textbf{Pre-training}  & \multicolumn{1}{c}{\textbf{Top 1}} & \multicolumn{1}{c}{\textbf{Top 5}} \\
		\midrule
		STC \cite{diba2018spatio} & ResNext101 & \xmark & 68.7 & 88.5 \\
		ARTNet \cite{wang2018appearance} & ResNet-18 &ImageNet & 69.2 & 88.3 \\
		S3D \cite{xie2018rethinking} & BN-Inception & ImageNet & 72.2 & 90.6 \\
		I3D \cite{carreira2017quo} & BN-Inception & ImageNet & 71.1 & 89.3 \\
		StNet \cite{he2019stnet} & ResNet-101 & ImageNet&71.4  & - \\
		Disentangling \cite{zhao2018recognize} & BN-Inception & ImageNet& 71.5  & 89.9 \\
		\midrule
		R(2+1)D \cite{tran2018closer} & ResNet-18 & \xmark & 56.8  & - \\
		R(2+1)D \cite{tran2018closer} & ResNet-34 & \xmark & 72.0  & 90.0 \\
		\midrule
		TSM \cite{lin2019tsm} & ResNet-50 & ImageNet &72.5  & 90.7 \\
		TSN  \cite{wang2016temporal} & BN-Inception & ImageNet& 69.1  & 88.7 \\
		STM \cite{jiang2019stm} & ResNet-50  & ImageNet & \textbf{73.7}  & \textbf{91.6} \\
		\midrule
		ST-STFT   & BN-Inception & \xmark  &	58.8 & 82.1\\
		S-STFT     & BN-Inception & \xmark & 57.3& 80.7\\
		T-STFT    & BN-Inception & \xmark & 61.1 & 85.2\\		
		\bottomrule
	\end{tabular}
	\label{tab:kinetics}
\end{table}

\section{Conclusion}\label{sec:conclusion}
In this paper, we propose a new class of layers called STFT that can be used as an alternative to the conventional 3D convolutional layer and its variants in 3D CNNs. The STFT blocks consist of a non-trainable depthwise 3D STFT micro-architecture that captures the spatially and/or temporally local information using STFT at multiple low frequency points, followed by a set of trainable linear weights for learning channel correlations. The STFT blocks significantly reduce the space-time complexity in 3D CNNs. Furthermore, they have significantly better feature learning capabilities. We demonstrate STFT-based networks achieve the state-of-the-art accuracy on temporally challenging and competitive performance on scene-related action recognition datasets. 
We believe the architectures and ideas discussed in this paper can be extended for tasks such as classification and segmentation for other 3D representations, such as 3D voxels and 3D MRI. We leave this as our future work.

\begin{table}[!t]
	\centering
	\caption{Performance of the X-STFT networks on the UCF-101 and HMDB-51 datasets compared with the state-of-the-art methods.}
	\begin{tabular}{lccc}  
		\toprule[.15em]
		\textbf{Model}    & \textbf{Pre-training} & \textbf{UCF-101} & \textbf{HMDB-51} \\
		\midrule
		I3D \cite{carreira2017quo} & ImageNet+Kinetics & 95.1 & \textbf{74.3}\\
		TSN \cite{wang2016temporal} & ImageNet+Kinetics & 91.1 & -\\ 
		TSM \cite{lin2019tsm} & ImageNet+Kinetics  & 94.5 & 70.7 \\
		StNet \cite{he2019stnet} & ImageNet+Kinetics & 93.5 & - \\
		Disentangling \cite{zhao2018recognize} & ImageNet+Kinetics & 95.9 & - \\
		STM \cite{jiang2019stm} & ImageNet+Kinetics & \textbf{96.2} & 72.2 \\
		C3D \cite{tran2015learning} &    Sports-1M &	82.3 & 51.6\\
		TSN \cite{wang2016temporal} & ImageNet & 86.2 & 54.7\\
		
		STC \cite{diba2018spatio} &  Kinetics & 93.7 & 66.8\\
		ARTNet-TSN \cite{wang2018appearance}  & Kinetics & 94.3  & 70.9 \\
		ECO \cite{zolfaghari2018eco} & Kinetics & 94.8 & 72.4 \\
		\midrule
		ST-STFT     &  Kinetics &	93.1 & 67.8\\
		S-STFT     & Kinetics &  92.2  & 66.6\\   
		T-STFT    &  Kinetics & 94.7 & 71.5\\
		\bottomrule[.15em]
	\end{tabular}
	\label{tab:ucf-hmdb}
\end{table}

\end{document}